# TSMS-SAM2: Multi-scale Temporal Sampling Augmentation and Memory-Splitting Pruning for Promptable Video Object Segmentation and Tracking in Surgical Scenarios

Guoping Xu[1], Hua-Chieh Shao[1], You Zhang[1#]

[1]The Medical Artificial Intelligence and Automation (MAIA) Laboratory

Department of Radiation Oncology

University of Texas Southwestern Medical Center, Dallas, TX 75390, USA

[#]Corresponding Email: You.Zhang@UTSouthwestern.edu

## Abstract

Promptable video object segmentation and tracking (VOST) has seen significant advances with the emergence of foundation models like Segment Anything Model 2 (SAM2); however, their application in surgical video analysis remains challenging due to complex motion dynamics and the redundancy of memory that impedes effective learning. In this work, we propose TSMS-SAM2, a novel framework that enhances promptable VOST in surgical videos by addressing challenges of rapid object motion and memory redundancy in SAM2. TSMS-SAM2 introduces two key strategies: multi-temporal-scale video sampling augmentation to improve robustness against motion variability, and a memory splitting and pruning mechanism that organizes and filters past frame features for more efficient and accurate segmentation. Evaluated on EndoVis2017 and EndoVis2018 datasets, TSMS-SAM2 achieved the highest mean (± s.d.) Dice scores of 95.24±0.96% and 86.73±15.46%, respectively, outperforming prior SAM-based and task-specific methods. Extensive ablation studies confirm the effectiveness of multiscale temporal augmentation and memory splitting, highlighting the framework's potential for robust, efficient segmentation in complex surgical scenarios. Our source code will be available at https://github.com/apple1986/TSMS-SAM2.

**Key words:** Video Object Tracking and Segmentation, Segment Anything 2, Video Augmentation, Memory

# 1. Introduction

Promptable video object segmentation and tracking (VOST) refers to the task where a user-provided prompt—such as a mask, point, or bounding box—is given on the first frame of a video, and the target object is subsequently tracked and segmented across frames. This is typically achieved by propagating memory based on preceding image features and prediction results [1, 2]. The field has seen significant advances following the introduction of foundation models like the Segment Anything Model (SAM) [3] and Segment Anything Model 2 (SAM2) [4], which are trained on large-scale natural image datasets. Notably, SAM2 incorporates a streaming memory mechanism within a transformer-based architecture, enabling more efficient and accurate VOST compared to its predecessor.

Despite these advancements, adapting such models to specialized domains like medicine remains challenging due to the domain gap between natural and medical imagery. As a result, fine-tuning or retraining on domain-specific datasets is often necessary. For instance, MedSAM2 [5] was developed by fine-tuning SAM2 using a medical dataset consisting of over 455,000 3D image-mask pairs and 76,000 video frames. In a related effort, Medical SAM2 [6] employed the One-Prompt dataset [7], a large-scale collection comprising 78 public medical datasets spanning various modalities, anatomical structures, and imaging conditions. These studies underscore the critical role of curated datasets in effectively transferring foundation models like SAM2 to medical applications, enabling strong generalization and robust segmentation for VOST. However, constructing large-scale medical video datasets remains a formidable challenge, as it demands extensive expert annotation and significant time investment due to the high frame counts involved in each sequence.

Image augmentation has been widely recognized as an effective and efficient strategy to mitigate the reliance on large amounts of labeled data in deep learning-based image segmentation [8]. Numerous augmentation techniques have been proposed and are now considered standard practice when training deep neural networks. These include basic geometric transformations (e.g., rotation, scaling, affine, and elastic deformations), intensity-based modifications (e.g., random blurring, sharpening, and noise addition), and more advanced compositional methods such as Copy-Paste [9, 10], RandAugment [11], Mixup [12], and CutMix [13].

However, these augmentation strategies are predominantly designed for static, image-level tasks and do not account for the temporal dependencies inherent in video data. Such frame-to-frame object motion is an essential component in VOST tasks. Recent research efforts have begun to explore augmentation strategies that preserve or exploit temporal consistency, acknowledging the importance of motion and sequence dynamics in improving model generalization and robustness in video-based applications [14-16]. Despite these advances, spatial-temporal augmentation methods often rely heavily on the characteristics of the original data, particularly assuming consistent frame rates and smooth motion patterns. This reliance can limit the model's ability to generalize to more challenging tracking scenarios—such as occlusions, abrupt object disappearances, or sudden changes in object speed—where the temporal continuity is disrupted. For instance, as shown in Figure 1 (a surgical video sequence), the large needle driver (in red box) exhibits rapid shifts between frames T=1 and T=2, followed by an abrupt disappearance between T=4 and T=5. Addressing the variations of temporal dynamics remains an open challenge for developing more resilient and adaptable video augmentation strategies, particularly for complex and domain-specific scenarios such as surgical scenes.

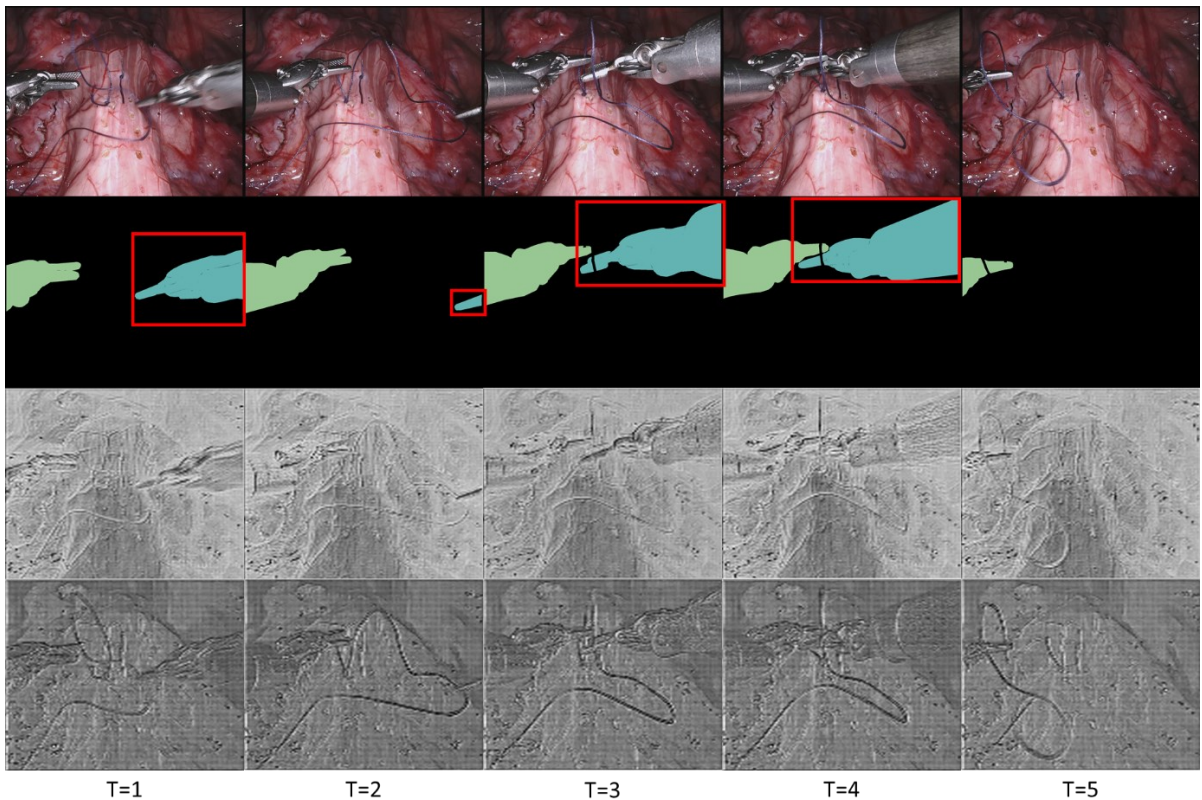

**Figure 1.** Illustration of substantial motion of a large needle driver (T=1 to T=2), which is marked in the red box. Abrupt object disappearances are also shown (T=4 to T=5). The first row shows the original video frames, and the second row presents the corresponding annotations. The third and fourth rows display the first and second channels of the feature maps extracted from the output of the first stage of SAM2's image encoder. For visualization purposes, all feature maps have been interpolated to match the resolution of the original input images.

In SlowFast networks [17], a dual-pathway architecture was introduced, consisting of a slow pathway and a fast pathway, designed to process the same video inputs at low and high frame rates, respectively. The slow pathway captures rich spatial semantics, while the fast pathway focuses on fine-grained motion dynamics through higher temporal resolution. This design has significantly improved the performance of action recognition and classification in videos, particularly for large-motion objects. Inspired by the SlowFast framework, we hypothesize that multi-scale temporal augmentation may offer an effective strategy for addressing two key challenges in VOST: the limited availability of annotated data and the variability in object motion speed. To this end, we explore the use of multi-rate temporal sampling as a data augmentation strategy during training. The core idea is to simulate a diverse range of motion patterns by varying temporal sampling rates, thereby enabling the model to better generalize across different motion dynamics.

In addition to time-based video sample augmentation, we also observed that the memory bank mechanism in SAM2, which sequentially stores features from previous frames, can introduce redundancy due to the inherent continuity of video data (see Figure 1, feature maps at timesteps T=3 and T=4). As demonstrated in SurgSAM2 [18], such redundancy not only increases computational overhead but may also hinder effective information propagation and timely feature updating for the current frame. To mitigate this, SurgSAM2 introduces a frame-pruning mechanism, which selectively retains the most informative past frame features to update the current frame representation. However, this approach primarily focuses on feature similarity with the latest frame in the memory bank, neglecting the distinct roles of short-term and long-term memory based on their temporal distance from the latest frame, both of which are crucial for effective video object segmentation [19]. Hence, it may inadvertently discard features that, while not immediately relevant, could be informative for future tracking or segmentation tasks. For example, in Figure 1, the features at T=3 and T=4 may be pruned due to lower similarity with T=5, despite their potential utility for future object tracking. These limitations underscore the need for more context-aware memory selection strategies that balance immediate relevance (short-term memory) with potential future utility (long-term memory) in VOST.

Hence, we propose two strategies to address the challenges associated with variable object motion speeds and redundant frame feature accumulation in SAM2 for VOST. The main contributions of this study are summarized as follows:

(1). We present a novel video augmentation method that leverages multiple temporal sampling rates to alleviate the scarcity of labeled video data and improve the model's robustness to rapid object motion.

(2). We introduce a new feature selection mechanism for SAM2 that partitions previous frame features into short-term and long-term memory, mitigating the limitations of over-reliance on the similarity with the latest frame in the memory bank for pruning frame features, and enhancing temporal context modeling.

(3). We evaluate the proposed methods on two publicly available surgical video datasets (EndoVis2017 and EndoVis2018). The experimental results demonstrate the effectiveness and generalizability of our approach in handling complex, real-world video segmentation scenarios.

## 2. Related work

### 2.1 Video augmentation

To alleviate the scarcity of video training samples and mitigate overfitting, several augmentation strategies have been specifically developed for video data, going beyond traditional static image-level techniques. For example, VideoMix [20] introduces a spatiotemporal augmentation approach in which random video cuboid patches are cut and pasted between different videos to enhance diversity for video classification tasks. In Learn2Augment [21], a selector module is designed to identify semantically similar video pairs, followed by a semantic matching and video compositing process that enables object-level cut-and-paste augmentation across videos with aligned contextual semantics, primarily for action recognition. Additionally, [22] proposes augmenting video at feature-level by injecting small Gaussian perturbations, aiming to reduce the effects of inter-video variance and model-induced noise in video relevance prediction. Most existing methods primarily focus on ensuring consistency or capturing semantic correlations within the spatiotemporal domain, and typically apply augmentation either at the data level or feature level [16, 23]. However, these augmentation strategies are generally confined to a single temporal scale, failing to account for the variations that occur across multiple temporal resolutions in object tracking and segmentation. This limitation may compromise robustness, particularly in challenging scenarios involving abrupt or rapid object motion.

### 2.2 Multi-scale temporal analysis of videos

Multi-scale (or multi-resolution) analysis has been widely applied in time-series signal processing to capture correlations across different temporal scales (frequencies) [24, 25]. In deep learning, previous studies on multi-temporal-scale video analysis have primarily focused on tasks such as video classification (e.g., action recognition [17, 26, 27]) and frame prediction [28, 29]. These approaches aim to capture both spatial semantics and temporal dynamics at various timescales (or frame rates) to facilitate a deeper contextual understanding of video content. For video classification, dual-stream architectures are commonly employed [26, 27], where one stream focuses on spatial appearance and the other on temporal motion cues. Multi-scale spatiotemporal features are typically extracted and fused through pooling operations. In addition, methods like SlowFast Networks [17] and Temporal Relation Networks [30] adopt multi-rate frame sampling strategies on the image level to enhance feature extraction for activity recognition and classification.

In contrast to these approaches, our work focuses on video sample augmentation via multi-frame-rate sampling strategies on videos to address the challenges posed by varying object motion speeds in SAM2-based VOST tasks. Rather than feature fusion, our method directly enriches the training data diversity, enhancing the model's robustness to object motion variations.

## 2.3 Memory management in SAM2-based methods

Features (memory) of previous video frames are critical for VOST as they provide essential recognition cues for the tracked objects. In SAM2, features from a fixed number of past frames are stored in a memory bank using a first-in-first-out (FIFO) updating mechanism. These stored features are uniformly fed into a memory attention module to condition the current frame's features. However, this design overlooks the potential redundancy among stored frame features—particularly as the number of stored frames increases—resulting in substantial computational overhead.

To address this, several subsequent studies have focused on improving the efficiency of memory management in SAM2. Two primary strategies have emerged for dynamically updating previous frame features: pruning-based and time-scale-based approaches. Representative pruning-based works include Medical SAM2 [6] and SurgSAM2 [18], which selectively prune redundant features in the memory bank based on confidence scores and feature similarity. For time-scale-based methods, notable examples include SAM-I2V [19] and MemorySAM [23]. In SAM-I2V, the original SAM is leveraged to extract a sequence of image features from video frames to serve as memory. A memory selective associator and a memory prompt generator are then introduced to manage the memory state and generate target object prompts based on the updated memory. Specifically, the memory is partitioned into local (short-term) and global frames (long-term) according to their temporal distance from the latest frame in the memory bank by the memory selective associator, which filters out redundant frame features. In MemorySAM, the previous feature frames are used to extract modality-agnostic information and memorize the semantics related to the object scene for multi-modality semantic image segmentation.

While these strategies have demonstrated effectiveness, most existing methods rely solely on the latest frame in the memory bank (the one that immediately precedes the current frame to be segmented/tracked) to guide feature pruning or memory grouping, overlooking the importance of long-term memory, which can be critical for handling challenging scenarios such as objects undergoing abrupt or large-scale motion, leading to the disappearance and re-appearance of objects. To address this gap, we propose to explicitly split the memory into short-term and long-term components, performing feature pruning within each part separately. This design aims to retain the most relevant features for the current frame while preserving potentially useful information in the long-term memory to benefit future frame updates.

## 3. Methods

In this study, we aim to enhance the model's robustness in handling rapid object motion through video sample augmentation and to improve inference efficiency/accuracy via memory pruning, building upon the powerful SAM2 framework. To this end, we propose TSMS-SAM2, a novel method that integrates multi-temporal-scale video sample augmentation and memory-splitting pruning. The overall architecture of the proposed TSMS-SAM2 is illustrated in Figure 2.

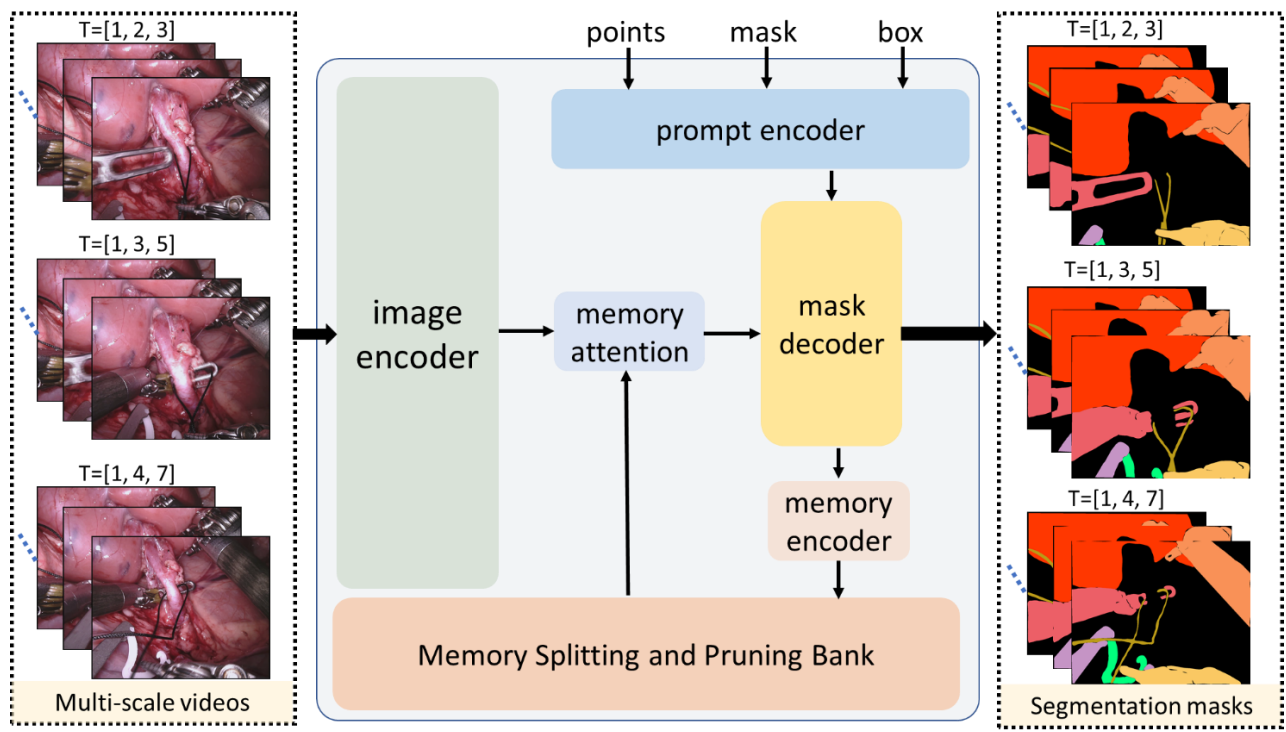

**Figure 2.** Overall architecture of the proposed TSMS-SAM2 framework. Compared to the vanilla SAM2, we introduce two key components: multi-scale temporal sampling for video augmentation and a memory splitting and pruning strategy. For the former, training videos are sampled at varying frame rates to augment objects with diverse motion speeds. For the latter, the memory bank is explicitly divided into short-term and long-term components, and previous feature frames are pruned separately to enhance both the efficiency and accuracy of SAM2.

### 3.1 SAM2's architecture

The SAM2 framework comprises six key components: an image encoder, a memory attention module, a prompt encoder, a mask decoder, a memory encoder, and a memory bank. The image encoder utilizes the Hiera model [31], pre-trained with Masked Autoencoders (MAEs) [32], to progressively extract multi-scale spatial features across stages. The memory attention module integrates previous frame features as memory via cross-attention to enhance the localization of tracked objects. The prompt encoder explicitly provides object position cues to the mask decoder, with the initial frame's mask commonly used as the prompt in promptable VOST tasks. Upon receiving updated current frame features through the attention module, the mask decoder outputs multiple masks corresponding to multiple objects per frame, and the masks are classified into different objects and passed to the memory encoder. The memory encoder fuses the selected mask features with corresponding image embeddings from the image encoder to extract memory information from the current frame. This information is then stored in the memory bank using a first-in-first-out (FIFO) mechanism to maintain a fixed number of previous frame features. While this streaming memory mechanism makes SAM2 well-suited for VOST, its fixed memory storage introduces redundancy, reduces efficiency, and may impair overall performance. To address these limitations, we propose TSMS-SAM2, which extends vanilla SAM2 by introducing a memory-splitting pruning strategy to dynamically select and retain the most informative features from previous frames.

### 3.2 Multi-temporal-scale video sampling augmentation

We propose a multi-temporal-scale video sampling augmentation strategy designed to enhance the model's robustness to varying object motion speeds, particularly in challenging scenarios involving occlusions, abrupt disappearances, or sudden motion shifts. The augmentation is formulated as:

$$V^S = \{I_t | t = t_0 + s \cdot k, k = 0,1,\cdots,K\} \tag{1}$$

Where $V^S$ denotes the sampled video sequence at the temporal scale $s$, $I_t$ is the video frame at time $t$, $t_0$ is the initial frame index, and $s \in S$ is a sampling stride selected from a predefined set of temporal scales $S = \{s_1, s_1, \cdots, s_n\}$. The hyperparameter $s$ controls the sampling rate, effectively generating multiple views of the video at different motion speeds. This augmentation encourages the model to learn motion representations that are resilient to temporal inconsistencies and distortions.

### 3.3 Memory splitting and pruning mechanism

The proposed memory splitting and pruning mechanism aims to remove redundant features from previous frames in the memory bank of SAM2 to improve computational efficiency/accuracy. This process consists of three key steps: (1) splitting the memory into short-term and long-term memory groups; (2) computing the feature similarity within each group; and (3) pruning redundant frames from each group and combining the retained features into the memory bank. The pipeline of memory splitting and pruning is illustrated in Figure 3.

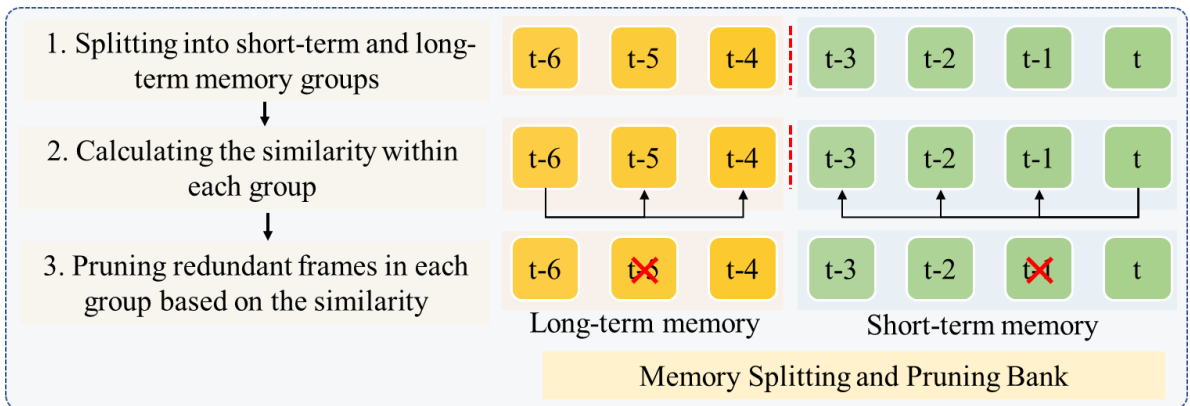

**Figure 3.** Illustration of the memory splitting and pruning process. The procedure consists of three main steps: (1) splitting the memory into short-term and long-term memory groups; (2) computing the similarity metric between the features of each stored frame and the features of a reference frame t-x within each group (x=6 for the long-term memory group and x=0 for the short-term memory group), excluding the reference frame itself; and (3) pruning the most redundant frame in each group based on the computed similarity scores. The retained frames from both groups are then combined and stored in the memory bank. The illustration is based on a maximum memory bank size of seven frames, the same as the original SAM2 framework.

Inspired by the Atkinson–Shiffrin memory model [33], which categorizes memory into sensory, short-term, and long-term components based on temporal duration, we adopt a temporal partitioning strategy for the memory bank. Specifically, frame-level features from previous time steps are divided into short-term and long-term memory groups according to their temporal distance from the latest frame $f_t$ in the memory bank, which immediately precedes the current frame under inference. Cosine similarity is then computed independently within each group to capture temporal relevance better and maintain discriminative representation.

For short-term memory, similarity is evaluated between the latest frame $f_t$ in the memory bank, and the other recent frames $\{f_{t-1}, f_{t-2}, \cdots, f_{t-(\lceil n/2 \rceil - 1)}\}$. For long-term memory, cosine similarity is computed between $f_{t-(n-1)}$ and the past frames $\{f_{t-\lceil n/2 \rceil}, \cdots, f_{t-(n-2)}\}$. Here, the $\lceil \cdot \rceil$ indicates the ceiling operation and $n$ indicates the maximum number of frames stored in the memory bank. The cosine similarity between two frame features $f_t$ and $f_i$ is defined as:

$$Similarity(f_t, f_i) = \sum_{c=1}^{N_c} \frac{f_t^{(c)} \cdot f_i^{(c)}}{\left\| f_t^{(c)} \right\| \left\| f_i^{(c)} \right\|} \tag{2}$$

where $c$ denotes the channel index out of a total of $N_c$ channels, and $t$ and $i$ refer to the temporal indices of the frame features.

To reduce redundancy in the memory bank, we identify and remove the frame feature with the highest similarity score within each group. The remaining features from both the short-term and long-term memory groups are then aggregated to construct the final memory bank for SAM2 to infer on the current frame.

## 4. Experiments

### 4.1 Implementation details

All experiments were performed on an Ubuntu system equipped with a single NVIDIA RTX 4090 GPU. The proposed TSMS-SAM2 model was developed using the SAM2-Small backbone. Following the configuration of SurgSAM2, we adopted an input resolution of 512 × 512, which aligns with practical dimensions in surgical imaging scenarios. During training, we fine-tuned the mask decoder, the memory attention module, and the memory encoder while keeping the image encoder and prompt encoder frozen to leverage the strong feature extraction capabilities of the pretrained SAM2 encoder. The model was trained for 30 epochs with a batch size of 2 using the AdamW optimizer and an initial learning rate of $5 \times 10^{-5}$. After inference, all predicted segmentations were resampled to the original video resolution for comparison with the 'ground truth'. Consistent with the vanilla SAM2 configuration, we maintained a memory bank of up to seven past frames to balance temporal context and computational efficiency. These frames were divided into two groups (up to four for short-term memory, and up to three for long-term memory). When the total past frames in the memory bank reached seven, the most similar frame in each group was pruned, resulting in five retained memory frames used for updating the current frame features. Otherwise, when the total past frames in the memory are fewer than seven, no pruning is performed.

### 4.2 Dataset

We evaluated our method on two widely adopted public surgical video segmentation benchmarks: the 2017 MICCAI EndoVis Instrument Segmentation Challenge (EndoVis17) [34] and the 2018 MICCAI EndoVis Scene Segmentation Challenge (EndoVis18) [35]. The EndoVis17 dataset consists of eight training sequences (225 frames each; 1,800 frames in total), eight subsequent testing sequences (75 frames each; 600 frames total), and two additional hold-out test sequences (sequences 9 and 10; 600 frames total), yielding a combined 3,000 annotated frames. This dataset includes six instrument classes: Bipolar Forceps, Prograsp Forceps, Large Needle Driver, Vessel Sealer, Grasping Retractor, and Monopolar Curved Scissors. The eight testing sequences are from the same procedures as the training sequences, although without any temporal overlap. To avoid potential data leakage, we exclusively used the two hold-out test sequences (sequences 9 and 10) comprising 600 frames to assess TSMS-SAM2, excluding the eight provided test sequences to ensure no overlap with training data and to better evaluate out-of-distribution performance.

The EndoVis18 dataset includes 15 videos, each containing 149 frames, resulting in a total of 2,235 annotated frames. It covers seven surgical instruments: Bipolar Forceps, Prograsp Forceps, Large Needle Driver, Monopolar Curved Scissors, Ultrasound Probe, Suction Instrument, and Clip Applier. Following the standard evaluation protocol established by ISINet [36] and SurgSAM2 [18], sequences 2, 5, 9, and 15 (596 frames) are used for testing, while the remaining 11 sequences (1,639 frames) are used for training.

### 4.3 Evaluation metrics

To comprehensively evaluate the video segmentation performance, we adopt several commonly used metrics in video/image object segmentation tasks:

(1) Jaccard Index (J or IoU): The Jaccard Index, commonly referred to as Intersection over Union (IoU), evaluates the similarity between the predicted segmentation P and the 'ground-truth' G, and is given by:

$$J = \frac{|P \cap G|}{|P \cup G|} \quad (3)$$

(2) Boundary F1 Score (F): This metric evaluates boundary alignment between predicted and 'ground-truth' segmentation masks. It is calculated by dilating boundary pixels of both masks with a disk of a 14-pixel radius in this study. Precision is the fraction of predicted boundary pixels within the dilated 'ground-truth' boundary, while recall is the fraction of 'ground-truth' boundary pixels within the dilated predicted boundary. Boundary F1 Score is the harmonic mean of this precision and recall, emphasizing contour accuracy.

$$F = \frac{2 \cdot Precision \cdot Recall}{Precision + Recall} \tag{4}$$

(3) J&F: This composite metric averages the regional similarity (Jaccard Index) and boundary accuracy (F1 Score) across frames. This metric provides a balanced assessment of both region-based and boundary-based segmentation quality.

$$J\&F = \frac{J + F}{2} \tag{5}$$

(4) Dice Similarly Coefficient (Dice): Dice measures the overlap between the predicted and 'ground-truth' masks, defined as:

$$Dice = \frac{2|P \cap G|}{|P| + |G|} \tag{6}$$

It ranges from 0 to 1, with higher values indicating better agreement. Dice is particularly sensitive to segmentation errors of small objects.

(5) Challenge IoU (CIoU): Adhering to the EndoVis2018 Challenge protocol, this metric assesses segmentation performance at the object level throughout the full video sequence. Rather than averaging IoUs frame by frame, CIoU calculates the intersection-over-union across the complete spatiotemporal region of each object.

$$CIoU = \frac{|\bigcup_{i=1}^{N} P_i \cap \bigcup_{i=1}^{N} G_i|}{|\bigcup_{i=1}^{N} P_i \cup \bigcup_{i=1}^{N} G_i|} \tag{5}$$

where $N$ is the total number of frames, and $P_i$, $G_i$ denote the predicted and 'ground-truth' masks in frame $i$, respectively. This metric accumulates the predicted and 'ground-truth' regions over all frames for each object and then computes a global IoU, ensuring a temporally consistent, object-centric evaluation of segmentation quality.

### 4.4 Experimental Results

*(1) Evaluation on the EndoVis2017 and EndoVis2018 datasets*

Table 1 presents a comprehensive performance comparison of our proposed TSMS-SAM2 against three baselines—SAM2, MedSAM2, and SurgSAM2—on the EndoVis2017 and EndoVis2018 datasets. Notably, SurgSAM2 previously achieved state-of-the-art performance on both datasets.

On EndoVis2017, TSMS-SAM2 achieves the highest performance across all four metrics: J, F, J&F, and Dice. Specifically, TSMS-SAM2 outperforms SurgSAM2 by approximately 0.93% (in absolute terms, same for the following) in J, 0.94% in F, 0.93% in J&F, and 0.95% in Dice, indicating the effectiveness of our multi-scale temporal sampling augmentation and memory pruning strategy. Compared to SAM2 and MedSAM2, TSMS-SAM2 achieves substantial gains of 5.89% and 7.87% in Dice, respectively, demonstrating enhanced segmentation accuracy under complex motion scenarios.

On the more challenging EndoVis2018 dataset, characterized by cluttered surgical scenes and domain shifts, TSMS-SAM2 again achieves the best overall performance. It surpasses SurgSAM2 by 5.05% in J, 5.59% in F, 5.31% in J&F, and 5.18% in Dice, clearly indicating its superior generalization capability. Notably, the performance margin over SAM2 and MedSAM2 is even more pronounced, with up to 14% improvement in Dice.

**Table 1**. Performance comparison on EndoVis2017 and EndoVis2018 datasets. The best values are shown in bold. The arrows are pointing in the direction of improved accuracy.

| Datasets | Method | J&F↑[%] | J↑[%] | F↑[%] | Dice↑[%] |
|---|---|---|---|---|---|
| EndoVis2017 | SAM2 | 86.69±7.39 | 85.45±8.00 | 87.93±6.96 | 89.35±8.27 |
| | MedSAM2 | 84.49±1.29 | 83.20±6.88 | 85.78±7.28 | 87.37±6.36 |
| | SurgSAM2 | 92.48±2.09 | 90.96±2.59 | 94.00±2.66 | 94.29±2.15 |
| | TSMS-SAM2 | **93.41±1.68** | **91.89±1.43** | **94.94±3.01** | **95.24±0.96** |
| EndoVis2018 | SAM2 | 69.81±22.76 | 69.37±22.78 | 70.25±22.95 | 72.79±23.77 |
| | MedSAM2 | 60.98±30.93 | 60.90±31.14 | 61.07±30.85 | 63.83±31.80 |
| | SurgSAM2 | 79.17±16.34 | 79.15±16.40 | 79.18±16.43 | 81.55±16.99 |
| | TSMS-SAM2 | **84.48±15.37** | **84.20±15.17** | **84.77±15.74** | **86.73±15.46** |

*(2) Ablation studies on the EndVis2018 dataset*

To further validate the effectiveness of our proposed components, we conduct ablation studies on the EndoVis2018 dataset, focusing on two key contributions: multi-temporal-scale sampling augmentation (TS) and memory splitting with pruning (MS). The results are summarized in Table 2.

When both TS and MS are disabled, the model shows the lowest performance across all four metrics, serving as the baseline. Introducing MS alone leads to a substantial improvement—approximately 8.1% in J&F and 8.3% in Dice—highlighting the importance of pruning redundant features for better memory utilization. Similarly, applying TS alone results in an increase of over 7% in J&F and Dice compared to the baseline, demonstrating that multi-scale temporal augmentation enables the model to better handle variations in motion speed and occlusion. Combining both TS and MS yields the best overall performance, with the highest scores in J&F (84.48%), J (84.20%), F (84.77%), and Dice (86.73%). These results confirm the complementary benefits of leveraging temporal diversity and memory structure refinement in enhancing segmentation robustness in complex surgical scenes.

Table 2. Ablation studies on TSMS-SAM2: with/without using multi-scale temporal sampling augmentation (TS) and memory-splitting pruning (MS). The best values are shown in bold. The arrows are pointing in the direction of improved accuracy.

| TS | MS | J&F↑[%] | J↑[%] | F↑[%] | Dice↑[%] |
|---|---|---|---|---|---|
| × | × | 76.28±25.00 | 76.12±24.67 | 76.43±25.41 | 78.44±25.22 |
| × | ✓ | 84.36±16.46 | 84.17±16.23 | 84.55±16.86 | 86.69±16.61 |
| ✓ | × | 83.13±20.84 | 82.87±20.74 | 83.40±21.07 | 85.35±21.14 |
| ✓ | ✓ | **84.48±15.37** | **84.20±15.17** | **84.77±15.74** | **86.73±15.46** |

Table 3 evaluates the effect of different temporal sampling rates used in the TS augmentation. The best performance across all metrics is achieved with the (1, 2) setting, where '1' denotes the original video frame rate and '2' indicates temporal down-sampling with a stride of 2. This setting is designed to generate training samples at multiple temporal scales, thereby improving robustness to variations in object motion. In contrast, using (1, 3) results in 6% Dice performance degradation, indicating that excessive temporal gaps harm feature alignment. The (1, 2, 3) setting offers moderate gains but remains inferior to the simpler (1, 2) configuration. These results suggest that a balanced and modest temporal stride improves segmentation effectiveness in surgical video.

Table 3. Ablation studies by using different sampling rate combinations for video augmentation. The best values are shown in bold. The arrows are pointing in the direction of improved accuracy.

| Sample frame rate scale | J&F↑[%] | J↑[%] | F↑[%] | Dice↑[%] |
|---|---|---|---|---|
| (1, 2)-used in this study | **84.48±15.37** | **84.20±15.17** | **84.77±15.74** | **86.73±15.46** |
| (1, 3) | 78.04±21.02 | 77.69±20.55 | 78.38±21.60 | 80.13±21.68 |
| (1, 2, 3) | 79.76±24.02 | 79.55±23.65 | 79.97±24.49 | 82.01±24.12 |

Table 4 compares different similarity metrics for pruning frame features in the memory bank. Cosine similarity yields the best performance, followed closely by Spearman correlation. Other metrics, such as L1, L2, dot product, and Pearson, perform significantly worse. These results highlight cosine similarity as the most effective and robust choice for memory pruning.

**Table 4**. Ablation studies on TSMS-SAM2: comparison between using different similarity measurements to prune frame features in the memory bank. The best values are shown in bold. The arrows are pointing in the direction of improved accuracy.

| Similarity | J&F↑[%] | J↑[%] | F↑[%] | Dice↑[%] |
|---|---|---|---|---|
| Cosine | **84.48±15.37** | **84.20±15.17** | **84.77±15.74** | **86.73±15.46** |
| Manhattan (L1) Distance | 81.43±16.98 | 81.19±16.90 | 81.68±17.20 | 83.77±17.31 |
| Euclidean Distance (L2) | 75.47±25.42 | 75.10±25.04 | 75.83±25.89 | 77.51±25.66 |
| Dot Product | 74.59±23.86 | 74.26±23.74 | 74.92±24.10 | 76.64±24.58 |
| Spearman Rank Correlation | 84.11±16.68 | 83.84±16.49 | 84.38±17.03 | 86.35±16.85 |
| Pearson Correlation | 78.35±28.48 | 78.14±28.47 | 78.56±28.60 | 80.68±28.97 |

*(3) Comparison of different methods on the EndoVis18 dataset with CIoU*

Table 5 compares different methods on the EndoVis2018 dataset using the Challenge IoU (CIoU) metric. Our proposed TSMS-SAM2 achieves the highest CIoU score of 85.1%, outperforming all previous task-specific and SAM-based methods.

Among task-specific approaches, MATIS Frame achieves the best performance at 82.4%, while SAM-based variants such as SurgSAM2 and MedSAM2 reach 84.4% and 83.7%, respectively. Earlier methods like TrackAnything and PerSAM fall significantly behind, especially under sparse point or mask inputs. These results demonstrate that TSMS-SAM2 leverages the generalization ability of SAM2 and significantly enhances the segmentation quality in complex surgical scenes through temporal modeling and memory optimization.

**Table 5**. Results of different methods on the EndoVis18 dataset. The best values are shown in bold. The arrows are pointing in the direction of improved accuracy.

| Method Category | Method | CIoU↑[%] |
|---|---|---|
| Task-specific | TernausNet [37] | 46.2 |
| | MF-TAPNet [38] | 67.9 |
| | ISINet [36] | 73.0 |
| | S3Net [39] | 76.2 |
| | MATIS Frame [40] | 82.4 |
| SAM-based | TrackAnything [41](1 Point) | 38.4 |
| | TrackAnything [41] (5 Points) | 60.9 |
| | PerSAM [42] (Mask) | 49.2 |
| | MaskTrack-RCNN [43] + SAM (bbox) | 78.5 |
| | Mask2Former [44] + SAM (bbox) | 78.7 |
| | SurgicalSAM [45] (bbox) | 80.3 |
| | SAM2 [4] (Mask) | 82.2 |
| | MedSAM2 [5] (Mask) | 83.7 |
| | SurgSAM2 [18] (Mask) | 84.4 |
| | **TSMS-SAM2 (Mask) (Ours)** | **85.1** |

*(4) Visual evaluation*

Figure 4 presents a qualitative comparison between our proposed TSMS-SAM2 and the baseline SurgSAM2 on the EndoVis2018 dataset. From top to bottom, the rows show the input image sequences, the corresponding 'ground-truth' masks, the predictions from SurgSAM2, and the predictions from TSMS-SAM2, respectively. Notably, SurgSAM2 fails to detect and segment the Large Needle Driver, whereas TSMS-SAM2 enables accurate and consistent tracking and segmentation. Figure 5 further shows a case where Prograsp Forceps and Ultrasound Probe are missed by SurgSAM2, while our TSMS-SAM2 effectively tracks and segments both instruments throughout the sequence.

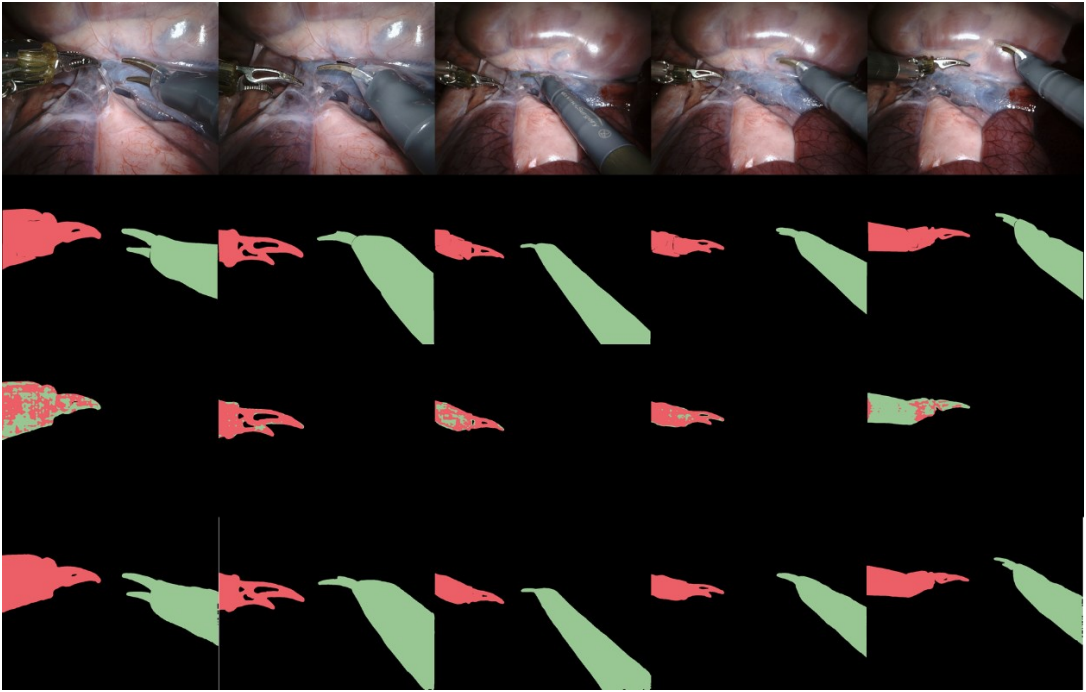

**Figure 4**. Qualitative comparison between TSMS-SAM2 and SurgSAM2 on the EndoVis2018 dataset. From top to bottom: (1) input image sequence, (2) 'ground-truth' segmentation, (3) predictions from SurgSAM2, and (4) predictions from TSMS-SAM.

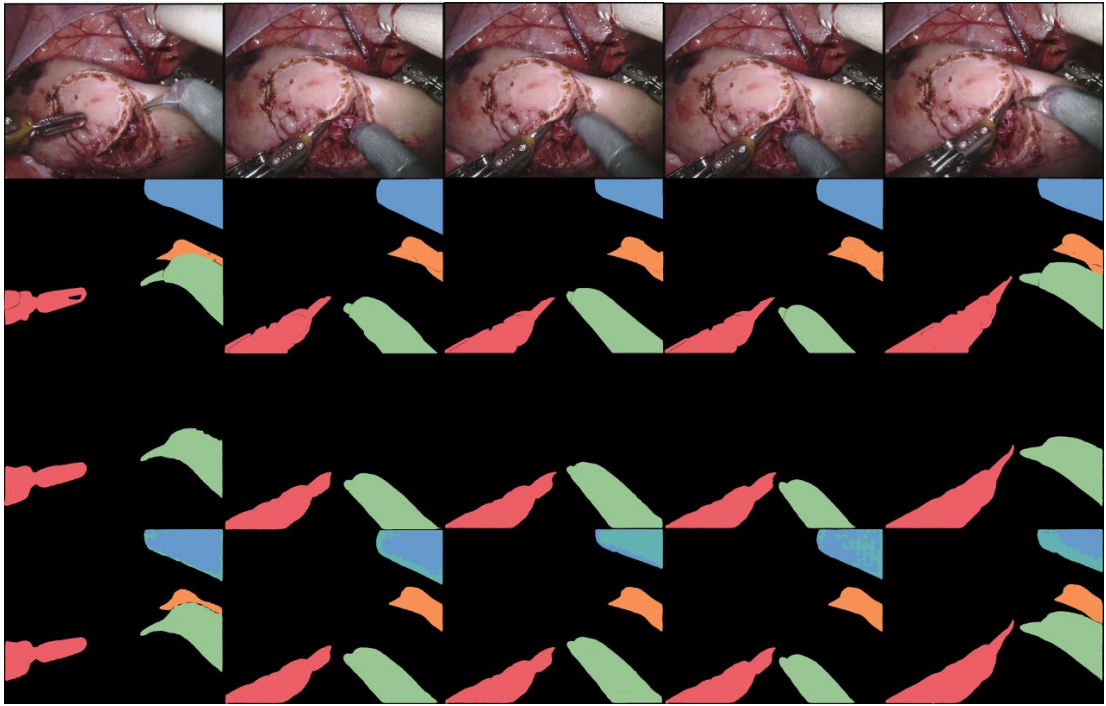

**Figure 5**. Visual comparison of instrument segmentation on the EndoVis2018 dataset. The rows illustrate: (1) input image sequence, (2) 'ground-truth' masks, (3) results from SurgSAM2, and (4) results from TSMS-SAM2.

## 5. Discussion and conclusion

In this study, we proposed two novel strategies—multi-scale temporal sampling for video augmentation and memory splitting with pruning—to enhance video object tracking and segmentation. Extensive experiments demonstrate the effectiveness of the proposed method. Nevertheless, there remain two main limitations that warrant future investigation.

*(1) Adaptive multi-scale temporal sampling for video augmentation*

The current implementation employs a fixed sampling rate (e.g., a rate of 1 or 2 in TSMS-SAM2) for temporal augmentation. While effective, this may not be optimal for all types of object motion. An adaptive sampling strategy, informed by scene dynamics or motion context, could better accommodate objects exhibiting rapid or irregular movement, thereby improving temporal consistency in challenging scenarios.

One potential solution is to leverage prior knowledge about object presence in each frame. For instance, if an object is absent in certain frames, lower sampling rates may suffice, whereas frames with continuous object presence may benefit from denser sampling. Furthermore, classical keypoint detection algorithms such as SIFT [46], FAST [47], or ORB [48] can be employed to extract motion-related cues from structural features across frames to drive the sampling strategy.

In addition, shape information can provide valuable guidance in determining sampling density. Fast-moving or highly deformable objects with elongated or irregular shapes may require more frequent temporal sampling to capture motion continuity, whereas compact or static shapes may permit sparser sampling. A similar principle is explored in RAFT [49], which utilizes dense optical flow to model motion dynamics across frames, demonstrating the benefit of motion-aware strategies in temporal processing.

*(2) Dynamic memory separation and memory updating*

We manually partition the memory bank into short-term and long-term components in this study. For example, with 7 stored frames, the most recent 4 are designated as short-term memory, and the remaining 3 as long-term memory. While simple, this static partitioning may not be optimal across diverse video sequences with varying temporal dynamics.

Advanced memory management strategies, such as those employed in XMem [50] and AOT [51], fuse and update memory features using dedicated modules designed to handle both short-term and long-term dependencies. One notable advantage of these approaches is their ability to maintain learned features over the entire video sequence, rather than being constrained to a fixed number of frames as in SAM2. However, this comes at the cost of increased computational overhead, as the whole memory needs to be recomputed and updated with every incoming frame, which can adversely affect the inference speed.

Future work could explore dynamic memory separation strategies, wherein a learnable model or rule-based mechanism adaptively partitions memory entries based on factors such as temporal distance, feature relevance, or similarity. This contrasts with our current approach, which employs a fixed division of short-term and long-term memory. Additionally, attention-based memory modules—similar to those used in XMem and AOT—could be integrated to selectively update short-term and long-term memories within the two memory groups after frame pruning, thereby improving memory efficiency without sacrificing performance.

In conclusion, we introduce a multi-scale temporal sampling strategy for data augmentation and a memory splitting framework for structured feature organization in SAM2. Evaluations on two publicly available surgical datasets confirm that our approach significantly improves segmentation performance, offering a promising direction for advancing video-based medical image analysis.

## Acknowledgement

The study was supported by the US National Institutes of Health (R01 CA240808, R01 CA258987, R01 EB034691, and R01 CA280135).